# Probabilistic inverse reinforcement learning in unknown environments

**Aristide C. Y. Tossou**
EPAC, Abomey-Calavi, Bénin
yedtoss@gmail.com

**Christos Dimitrakakis**
EPFL, Lausanne, Switzerland
christos.dimitrakakis@gmail.com

## Abstract

We consider the problem of learning by demonstration from agents acting in unknown stochastic Markov environments or games. Our aim is to estimate agent preferences in order to construct improved policies for the same task that the agents are trying to solve. To do so, we extend previous probabilistic approaches for inverse reinforcement learning in known MDPs to the case of *unknown* dynamics or opponents. We do this by deriving two simplified probabilistic models of the demonstrator's policy and utility. For tractability, we use maximum a posteriori estimation rather than full Bayesian inference. Under a flat prior, this results in a convex optimisation problem. We find that the resulting algorithms are highly competitive against a variety of other methods for inverse reinforcement learning that do have knowledge of the dynamics.

## 1 Introduction

We consider the problem of learning by demonstration from agents acting in stochastic Markov environments, or in stochastic Markov games [13, 18], when we do not know the underlying dynamics of the environment or the opponent strategy. This type of learning is very useful, since it can significantly decrease the time needed to acquire a particular task. Another possible application are games such as chess, where large libraries of expert play have been accumulated over the years. Inverse reinforcement learning offers a principled method to use this data for imitating expert play, even when the experts are deviating from the optimal strategy. In this paper, we learn through demonstration by estimating the preferences and policy of the agent giving the demonstration. These can then be used to estimate improved policies, by combining the inferred agent preferences with policy improvement schemes. This is not always trivial, since the demonstrations may be sub-optimal and the underlying dynamics are unknown to us.

Our main *technical contribution* is two inverse reinforcement learning algorithms for the case of *unknown dynamics*. These are inspired by two Bayesian models for inverse reinforcement learning in known environments [7, 17]. Our first algorithm extends the original model to unknown dynamics by coupling probabilistic preference and policy estimation with approximate dynamic programming schemes and maximum a posteriori estimation. Thus, we avoid explicitly estimating a model altogether. Our second algorithm proposes a simpler model where the policy and value function are jointly represented by the same set of parameters. We can thus eliminate both the dynamic programming step and estimation of the dynamics. Finally, we *experimentally* evaluate our schemes against a broad selection of algorithms which *do* know the dynamics. We consider both agents acting in unknown environments, as well agents playing stochastic games, and show that we can approach and even surpass the performance of methods that use knowledge of the dynamics.

The remainder of the paper is organised as follows. Section 2 formally introduces the setting, while Section 3 discusses related work and our contribution. Section 4 describes our models and algorithms. Comparisons with other methods are given in Section 5, and we conclude with a discussion in Section 6.

## 2 Setting

In our setting, we observe a set of demonstrations $D$ from an agent acting in an unknown Markov environment or game $\nu \in \mathcal{N}$, where $\mathcal{N}$ is a set of possible environments. The $k$-th demonstration $d_k \in D$ is a $T_k$-long trajectory, consisting of a sequence of environment states $s \in \mathcal{S}$ and agent actions $a \in \mathcal{A}$ with

$d_k = ((s_1^k, a_1^k), \ldots, (s_{T_k}^k, a_{T_k}^k))$. We assume that both the agent policy $\pi$ and the environment $\nu$ are Markovian. Consequently, the action $a_t$ only depends on the current state $s_t$ and the next state $s_{t+1}$ only depends on $s_t, a_t$. The agent acts according to some (unknown to us) reward function $\rho : \mathcal{S} \times \mathcal{A}$. In particular, any policy $\pi$ that the agent chooses, has a value defined in terms of the expected utility with respect to $\rho$:

$$V_{\nu,\rho}^\pi(s) \triangleq \mathbb{E}_{\pi,\nu}(U_\rho \mid s_1 = s), \quad U_\rho \triangleq \sum_{t=1}^T \rho(s_t, a_t), \tag{2.1}$$

where $T$ may be random.[1] We denote the optimal policy for a reward function by $\pi^*(\nu, \rho)$, with either of those two arguments omitted if it is clear from context. The optimal value function is denoted by:

$$V_{\nu,\rho}^* \triangleq \sup_\pi V_{\nu,\rho}^\pi. \tag{2.2}$$

Similarly, we define $Q_{\nu,\rho}^\pi(s, a) \triangleq \mathbb{E}_{\pi,\nu}(U_\rho \mid s_1 = s, a_1 = a)$ to be the expected utility of taking action $a$ at state $s$ and following $\pi$ thereafter, while $Q_{\nu,\rho}^*(s, a)$ is the optimal $Q$-function. We assume that the agent's policy is $\varepsilon$-optimal with respect to $\rho$, that is:

$$V_{\nu,\rho}^\pi \geq V_{\nu,\rho}^* - \varepsilon. \tag{2.3}$$

In our setting, we can observe the sequence of states and actions taken and we know (the distribution of) $T$. However, the policy $\pi$, the environment $\nu$ and the reward function $\rho$, as well as the $\varepsilon$-optimality of the policy are *unknown* to us.

## 3 Related work and contribution

Inverse reinforcement learning [15] is the problem of estimating the reward function of an agent acting in a dynamic environment. It is thus closely related to preference elicitation [3], since the reward function can be used to calculate the agent preferences, and apprenticeship learning [1], since for each reward function one can calculate the optimal policy.

Much previous work in this setting employs a linear approximation with feature expectations. The general idea is to find a reward function minimising some loss between the expert features expectations and the one of the optimal policy for the approximated dynamics and reward function. A representative example is [8], which uses the quadratic programming maximum-margin approach proposed in [1] to obtain a reward function and least squares temporal differences

[1]For geometrically distributed $T$, the expected utility function is identical to that obtained when $T \to \infty$ and the rewards are exponentially discounted.

(LSTD) [4] for policy evaluation. LSTD is also used to estimate the feature expectation of the expert. A similar approach is taken in [14], which used the game-theoretic algorithm MWAL [20] where a near optimal policy is found by using LSPIf, a variant of LSPI [11] for discrete finite horizon Markov decision processes.

Other methods avoid using policy iteration, but nevertheless indirectly employ knowledge of the environment. The most important work in this category is relative entropy inverse reinforcement learning [2]. The main idea is to perform a sub-gradient ascent on an appropriate loss. The sub-gradient is estimated using importance sampling on trajectories generated using a stochastic policy on the real environment. Another work in this category is the SCIRL (Structured Classification for Inverse Reinforcement Learning) algorithm [9]. Using the observation that the reward and the $Q$ function share the same parameter, then estimate it by minimising an appropriately defined loss. While we use a similar idea, SCIRL uses the real environment model to calculate feature expectations.

An altogether different approach is taken by [12], which constructs reward features from a set of component features through logical conjunctions. Given a set of trajectories, and the true environment model, the algorithm then estimates a reward function by finding the most effective feature combinations.

Our algorithms are inspired by two recently proposed probabilistic models for Bayesian inverse reinforcement learning. The first model starts by specifying a prior on the reward function and a reward-conditional distribution on policies through a prior on policy randomness [17]. The second model instead specifies a prior on the policy directly and policy-conditional distribution on the reward functions through a prior on policy optimality [7]. These methods have been described and evaluated on problems where the dynamics are known and the authors suggest that handling the unknown dynamics case would be possible by simply adding a prior distribution on the dynamics.

### 3.1 Our contribution

Our own work focuses on the problem of learning from demonstration in the case where the transition model of the process is *unknown.* This is the case when the agent is either acting in an environment with unknown dynamics, or playing an alternating Markov game against an unknown opponent. Unlike most previous approaches, we do not have prior knowledge of the dynamics, either in the form of an analytically available transition kernel, or in the form of a simulator from which we can take samples.

With respect to the reward-prior model specified in [17], our main technical novelty is to avoid estimating the dynamics altogether. Instead, we use *approximate dynamic programming* (LSTDQ) in combination with linear parametrisation to obtain policies and value functions for the reward prior model.

With respect to the policy-optimality-prior specified in [7], our main technical novelty is that we *eliminate the need* for a dynamic programming step altogether. This significant simplification is achieved by placing a prior on value functions rather than reward functions. Then, instead of considering all possible value functions for which a given policy is $\varepsilon$-optimal, we restrict the space by jointly parametrising the policy and value function.

Thirdly, our models use *features*, rather than the raw state observations. Thus our algorithms are *more generally applicable*.

Finally, for *computational simplicity*, we employ maximum a posteriori (MAP) estimation, rather than full Bayesian inference, as also suggested in [5]. Then inference can be performed quickly using any appropriate global or local optimisation algorithm.

Our main *experimental* contribution is to show that our proposed algorithms are *highly competitive* against six other methods, even if those are using *knowledge* of the dynamics. In particular, we investigate the robustness of our algorithms, by tuning the hyper-parameters, including the features, on a fixed set of demonstrations and then testing their performance on a large sample of demonstrations drawn from some distribution.

We compare our approaches with the full Bayesian method RPB in [17], the MAP$_{\text{IRL}}$ method in [5], MWAL [20], the maximum entropy method MAX-ENT [21], the projection method PROJ suggested in [1], MMP [16] and finally the feature-based FIRL [12]. We outperform most of those methods in all of our experiments, apart from the FIRL which (naturally) had particularly good performance in a domain with a high number of features. However, even in this case, our algorithms, without any knowledge of the dynamics, manage to approach the FIRL solution.

## 4 Models

The algorithms we provide are inspired by the probabilistic models introduced in [7, 17], whose difference from our models we describe in this section. Both of these maintain a joint prior distribution $\xi$ on policies $\pi \in \mathcal{P}$ and reward functions $\rho \in \mathcal{R}$. They only differ on how they model their dependencies. The *reward prior* model specifies a prior on reward functions and a conditional distribution on policies given the reward.

The *policy optimality* model specifies a prior on policies and a conditional distribution on reward functions given the policy. In both cases, the likelihood is simply the probability $p(D \mid \pi)$ of observing the demonstrations under the estimated policy.

Both of our models use features, rather than direct state observations. Thus, we define a mapping $g_R : \mathcal{S} \times \mathcal{A} \to \mathcal{X}_R$ from the state-action space to a feature space $\mathcal{X}_R \subset \mathbb{R}^{m_R}$ for the reward function. Similarly, we also define a mapping $g_Q : \mathcal{S} \times \mathcal{A} \to \mathcal{X}_Q$ from the state-action space to a feature space $\mathcal{X}_Q \subset \mathbb{R}^{m_Q}$ for the state-action value function.

### 4.1 Reward prior (RP) model

The main difference between the original model and ours is with respect to the reward function parametrisation and the value function estimation. In our model, we maintain a parameter $\boldsymbol{w}_R \in \mathbb{R}^{m_R}$ parametrising the reward function via a linear function $f : \mathcal{X}_R \times \mathbb{R}^{m_R} \to \mathbb{R}$, such that:

$$\rho(s,a) \triangleq f(g_R(s,a), \boldsymbol{w}_R) \triangleq g_R(s,a)^\top \boldsymbol{w}_R. \quad (4.1)$$

Since the environment dynamics are not known, we employ least-square temporal differences (LSTDQ [11]) to obtain a parametrised state-action value function:

$$\hat{Q}^*(s,a) = g_Q(s,a)^\top \boldsymbol{w}_Q. \quad (4.2)$$

The use of LSTDQ allows us to consider the case where the demonstration $D$ comes from a suboptimal expert. Since $D$ is generated by the expert's policy, we can use the on-policy LSTDQ. This finds a parameter $\boldsymbol{w}_Q$ such that $\boldsymbol{w}_Q = A^{-1}b = A^{-1}Z\boldsymbol{w}_R = C\boldsymbol{w}_R$, with $C = A^{-1}Z$ and :

$$A = \sum_{d_k \in \mathcal{D}} \sum_{t=1}^{T_k-1} g_Q(s_t^k, a_t^k) \left(g_Q(s_t^k, a_t^k) - \gamma g_Q(s_{t+1}^k, a_{t+1}^k)\right)^\top \quad (4.3)$$

$$b = \sum_{d_k \in \mathcal{D}} \sum_{t=1}^{T_k-1} g_Q(s_t^k, a_t^k) \rho(s_t^k, a_t^k) \quad (4.4)$$

$$= \sum_{d_k \in \mathcal{D}} \sum_{t=1}^{T_k-1} g_Q(s_t^k, a_t^k) g_R(s_t^k, a_t^k)^\top \boldsymbol{w}_R \quad (4.5)$$

$$= \left(\sum_{d_k \in \mathcal{D}} \sum_{t=1}^{T_k-1} g_Q(s_t^k, a_t^k) g_R(s_t^k, a_t^k)^\top\right) \boldsymbol{w}_R \quad (4.6)$$

$$= Z\boldsymbol{w}_R. \quad (4.7)$$

So:

$$Z = \sum_{d_k \in \mathcal{D}} \sum_{t=1}^{T_k-1} g_Q(s_t^k, a_t^k) g_R(s_t^k, a_t^k)^\top. \quad (4.8)$$

As the demonstration $D$ is fixed, so are $A$, $Z$ and $C$. Then,
$$\hat{Q}^*_{\boldsymbol{w_R}}(s,a) = g_Q(s,a)^\top C \boldsymbol{w_R}. \qquad (4.9)$$

As in the original model, we assume that, the demonstration policy $\pi$ is *softmax* with respect to the value function:
$$\pi_{\beta, \boldsymbol{w_R}}(a \mid s) \triangleq \frac{e^{\beta \hat{Q}^*_{\boldsymbol{w_R}}(s,a)}}{\sum_{a'} e^{\beta \hat{Q}^*_{\boldsymbol{w_R}}(s,a')}}, \qquad (4.10)$$

Finally, the likelihood is simply:
$$p(D \mid \beta, \boldsymbol{w_R}) = \prod_{d_k \in \mathcal{D}} \prod_{t=1}^{T_k} \pi_{\beta, \boldsymbol{w_R}}(a_t^k \mid s_t^k), \qquad (4.11)$$

and the overall posterior can be factorised due to conditional independence:
$$\xi(\beta, \boldsymbol{w_R} \mid D) \propto p(D \mid \beta, \boldsymbol{w_R}) \xi(\boldsymbol{w_R}) \xi(\beta). \qquad (4.12)$$

The main question is which prior to select. If we assume a uniform prior for the reward parameters $\boldsymbol{w_R}$ and $\beta$, taking the logarithm and replacing $\hat{Q}^*_{\boldsymbol{w_R}}$ by its value results in the following maximisation problem:

$$\max_{\boldsymbol{w_R}} \sum_{d_k \in \mathcal{D}} \left\{ \sum_{t=1}^{T_k} \beta g_Q(s_t^k, a_t^k)^\top C \boldsymbol{w_R} - \ln \sum_{a'} e^{\beta g_Q(s_t^k, a')^\top C \boldsymbol{w_R}} \right\}. \qquad (4.13)$$

Equation 4.13 is a concave function which can be maximised efficiently. Note that since $\boldsymbol{w_R}$ is not constrained and the prior for $\boldsymbol{w_R}, \beta$ is uniform, $\beta$ is essentially a free parameter which can be set to an arbitrary positive value. Then, we can get both the preference of the expert $\boldsymbol{w_R}$, as well as the parameter of the value function $\boldsymbol{w_Q}$.

An *alternative* is to use an exponential prior for $\beta$ and a Dirichlet prior for $\rho$. Intuitively, this should induce a penalty for overfitting, and a sparse reward function. However, the optimisation is in this case difficult, and in preliminary experiments we found that it did not improve upon the uniform prior.

## 4.2 Policy optimality (PO) model

This model starts with a prior $\xi(\pi)$ on the policies. This allows a posterior on the policy $\xi(\pi \mid D)$ to be obtained from the data directly. This type of "imitator" policy is then tempered through use of a prior on *policy optimality*. Briefly, if the policy is $\varepsilon$-optimal, then there is a set of reward functions $\mathcal{R}_\varepsilon$ such that $V^\pi_{\nu,\rho} \geq V^*_{\nu,\rho} - \varepsilon$ for any $\rho \in \mathcal{R}_\varepsilon$.

The *original* model assumed a prior $\xi(\varepsilon) = e^{-\varepsilon}$, for $\varepsilon$-optimality, and so obtained a distribution on reward functions conditional on the policy and its optimality. The overall posterior was then factorised as follows:

$$\xi(\varepsilon, \pi, \rho \mid D, \nu) \propto p(D \mid \pi, \nu) \xi(\rho \mid \pi, \varepsilon) \xi(\varepsilon) \xi(\pi). \qquad (4.14)$$

The difficulty with this model is that one has to consider all reward functions for which a policy is $\varepsilon$-optimal. This is not a problem for a finite set reward functions, but it makes inference hard in the general case. In addition, since $\nu$ is not known, one would have to integrate (or maximise) over $\nu$ explicitly.

Our own model considers *value functions* directly, rather than reward functions, thus eliminating the need for a dynamic programming step. In particular, a vector $(\boldsymbol{w_Q}, \beta)$ with $\boldsymbol{w_Q} \in \mathbb{R}^{m_Q}$, jointly parametrises the optimal value function and the demonstrator policy. Specifically, the optimal state-action value function is parametrised via a linear function $h : \mathcal{X}_Q \times \mathbb{R}^{m_Q} \to \mathbb{R}$, such that:

$$Q(s,a) = h(g_Q(s,a), \boldsymbol{w_Q}), \qquad (4.15)$$

while the policy is defined as:

$$\pi_{\beta, \boldsymbol{w_Q}}(a \mid s) = \frac{e^{\beta Q(s,a)}}{\sum_{a' \in \mathcal{A}} e^{\beta Q(s,a')}}. \qquad (4.16)$$

Each parameter $\beta, \boldsymbol{w_Q}$ corresponds to a unique policy-value function pair. We assume an independent prior $\xi(\beta, \boldsymbol{w_Q}) = \xi(\beta) \xi(\boldsymbol{w_Q})$. Our posterior probability is then:

$$\xi(\beta, \boldsymbol{w_Q} \mid D) \propto p(D \mid \pi_{\beta, \boldsymbol{w_Q}}) \xi(\beta) \xi(\boldsymbol{w_Q}), \qquad (4.17)$$

since given $\beta, \boldsymbol{w_Q}$, the policy and value function are uniquely determined.

Using uniform priors for all parameters, and taking the logarithm, results in the following maximisation problem:

$$\max_{\boldsymbol{w_Q}} \sum_{d_k \in \mathcal{D}} \left\{ \sum_{t=1}^{T_k} \beta g_Q(s_t^k, a_t^k)^\top \boldsymbol{w_Q} - \ln \sum_{a'} e^{\beta g_Q(s_t^k, a')^\top \boldsymbol{w_Q}} \right\}. \qquad (4.18)$$

It is easy to see that PO model can also be obtained by replacing $C \boldsymbol{w_R}$ in equation 4.13 by the value $\boldsymbol{w_Q}$. Although both PO and RP look similar, the PO model does not require the matrix $C$. Intuitively we can see that PO considers value functions directly, rather than reward functions, thus eliminating the need for a dynamic programming step. In particular, a vector $\boldsymbol{w_Q}$ jointly parametrises the optimal value function and the demonstrator policy defined in equation 4.10. Another significant difference is that PO starts with a prior $\xi(\pi)$ on the policies. This allows a posterior on the policy $\xi(\pi \mid D)$ to be obtained from the data directly.

An *alternative model* is to use an exponential distribution, $\xi(\beta^{-1}) = \mathcal{E}xp(1)$ for the scaling parameter, and a Dirichlet distribution $\xi(\boldsymbol{w_Q}) = \mathcal{D}irichlet(\boldsymbol{\alpha})$ for the other value function parameters, with $\boldsymbol{\alpha} \in \mathbb{R}_+^{m_Q}$ such that $\alpha_i = \frac{1}{2}$, so the posterior is:

$$\xi(\beta, \boldsymbol{w_Q} \mid D) \propto \prod_t \pi_{\beta,\boldsymbol{w_Q}}(a_t \mid s_t) \prod_{i=1}^{m_Q} w_{Qi}^{\alpha_i - 1} e^{-\beta^{-1}}. \tag{4.19}$$

Taking the logarithm, we now need to solve the following maximisation problem, under the constraint $\|\boldsymbol{w_Q}\|_1 = 1$, and $w_{Qi} \in [0, 1]$:

$$\max_{\beta,\boldsymbol{w_Q}} \sum_t \ln \pi_{\beta,\boldsymbol{w_Q}}(a_t \mid s_t) - \frac{1}{2} \sum_{i=1}^{m_Q} \ln w_{Qi} - \beta^{-1}. \tag{4.20}$$

Now $\beta^{-1}$ can be seen as a simple penalty to avoid overfitting. Due to the constraints, parameters $\boldsymbol{w_Q}$ where a few components have large values are favoured. Thus, this model can be seen as a sparse regularised classifier.

## 5 Experiments

We performed a set of experiments in three domains. These are either stochastic environments or alternating Markov games. In all those experiments, our algorithms did not assume knowledge of the underlying process, or opponent. For those cases where it was possible to obtain an *analytic* model of the process and/or of the opponent, we also compared our algorithms with methods which assume knowledge of the underlying dynamics. In all cases, we first tuned hyper-parameters of the algorithms in a small set of demonstration trajectories. These algorithms were then evaluated on multiple runs with varying amounts of demonstration trajectories, in order to examine the robustness of algorithms and their performance improvement as the amount of data increased.

As mentioned in the previous section, we have a choice of different priors for the models. The uniform priors lead to convex problems, which allow us to use an efficient algorithm (L-BFGS) to find the maximum a posteriori parameters. In preliminary experiments with the alternative priors, we had to resort to random restarts combined with local search. This didn't led to better results, possibly due to the difficulty of the optimisation problem. Consequently, in the presented experiments, we only show results with uniform priors. We use LPO to denote the linear PO model and LRP for the linear RP model.

All the experiments are performed on discrete environments, even though inference is performed on a set of features instead. Consequently, it is possible to compute the optimal policy with respect to a given reward function on all of these environments. In our experiments, we compare the performance of the policies found by the inverse RL algorithms to that of the optimal policy of the environment. In particular, the loss of a policy $\pi$ is:

$$\ell(\pi) \triangleq \sum_{s \in \mathcal{S}} \mu(s)(V^*(s) - V^\pi(s)), \tag{5.1}$$

where $\mu$ is the starting state distribution of the problem.

In all these experiments, the features are appropriately scaled for each algorithm. For example, when using MWAL the features are scaled so that they lie in $[-1, 1]$, whereas for PROJ, they lie in $[0, 1]$. For all experiments the features of the reward function $g_R$ are state features.

### 5.1 Blackjack

The first domain is blackjack, as described in [19]. The goal is to get cards such that their sum is as close as possible to 21 without exceeding it. All face cards count as 10 and the ace can be either 1 or 11 (*usable*). At the beginning of the game, two cards are dealt to both the dealer and the player. One of the card of the dealer is face up, the other is face down. If the player has immediately 21 (a face and an ace), then it is a *natural* and he wins (+1.5 points). Otherwise, he can request additional cards one by one (*hits*) until he either stops (*sticks*) or exceeds 21 (*goes bust*). If he goes bust, he loses (−1 points); if he sticks, then it is the dealer's turn. The dealer sticks when he has 17 or greater; he hits otherwise. If the dealer goes bust, then the player wins (+1 points); otherwise the outcome is determined by who ever is closer to 21.

The state is determined by the sum of the player's hand (12-21), the dealer card (ace-10 or 1-10), and whether or not the player holds a usable ace. We ignore the case where the player sum is less than 12, as then the player will always hit. This results in 201 states, including the terminal state.

We used 14 features for the reward function $g_R$. They included the state variables, their higher order (multivariate) polynomial terms (up to 2) and a bias (which is always 1). For the value function features $g_Q$, we repeated the reward features for each action. In this domain, the initial state distribution $\mu$ is derived from the uniform distribution on the set of cards.

The demonstrations are obtained from the optimal policy. We varied the number of episodes (from 10 to 10000) to determine the ability of the different algorithms to handle limited data. We report average performance on 200 runs. This includes both the loss (5.1) and CPU time.

## 5.2 Gridworld

The second experiment is on $32 \times 32$ gridworlds. The agent can move into 5 directions (west, east, north, south, or be still). But with probability 0.3 it fails and moves to a random direction. The grids are partitioned into 3 non overlapping regions (one at the lower left corner, one at the upper right corner and the third elsewhere). The initial state is uniformly drawn from the possible states. The true reward is positive in the two regions at the corner and negative elsewhere.

We used 64 features for the reward function $g_R$. These features included the coordinates $x$ and $y$ and 62 binary features indicating if $x$ or $y$ is lower than some value. For the value function features $g_Q$, we repeated the reward features for each action.

The demonstrations are obtained from an optimal agent. The number of steps for each episode is 8. The number of episodes is varied from 10 to 10000. We measured the average performance loss with respect to the optimal value function, over 10 experimental runs.

## 5.3 Tic-Tac-Toe

The final experiment is the game of tic-tac-toe. It is a $3 \times 3$ board game where two players (X and O) take turns alternatively to place their mark on empty spaces. The player who first places three marks in a row, horizontally, vertically, or diagonally wins and obtains a reward of $+1$. If there is no legal move remaining, the game is a draw. The game state can be described by 9 factors, each indicating the mark at a board location. Thus the total number of states is at most $3^9$, symmetry and unreachable states notwithstanding. There are up to 9 possible actions.

To construct $g_R$, we used the features described in [10]: the number of singlets (horizontal, diagonal, vertical lines with exactly one symbol X or O), doublets (horizontal, diagonal, vertical lines with exactly two symbols X or O), triplets (horizontal, diagonal, vertical lines with exactly three symbols X or O), diversity (the number of different singlet directions for each player) and crosspoints (an empty field belonging to at least two singlets of the same player) for each player; their

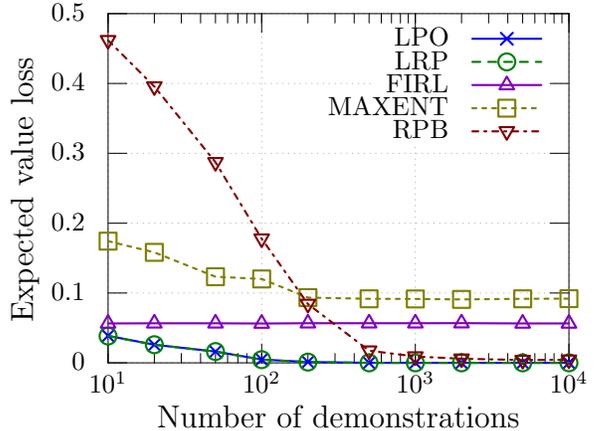

(a) Loss (5.1)

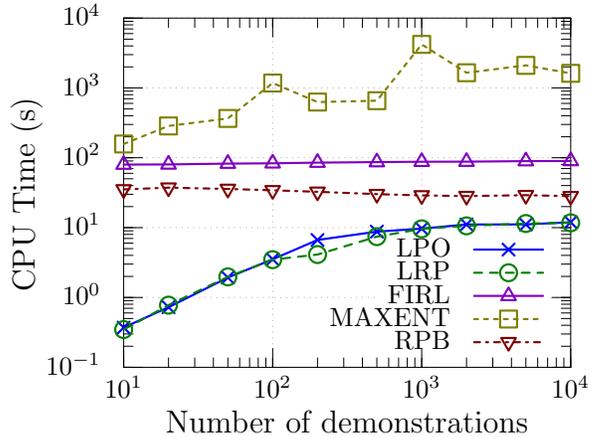

(b) CPU time

Figure 1: Blackjack results.

higher order (multivariate) polynomial terms (up to 2) as well as the 9 features for the raw board position.

The features $g_Q$ of the value function for a state-action pair $(s, a)$ are those of the resulting state $s'$ prior to the opponents play. We learned the policy for player X. The demonstrations are obtained from a game between the optimal player (X) and a uniformly random player (O). Player X randomly selects one of the available minimax-optimal actions. We performed the experiments by varying the number of demonstrations (from 10 to 10000) and report results averaged over 200 runs. We evaluated two cases: one against a random player and one against an optimal player.

## 5.4 Results

For clarity of presentation, the figures only compare our algorithms with the original reward prior model RPB, the maximum entropy method MAX-ENT [21] and the feature-based FIRL [12]. While results for

MWAL [20], MAP$_{IRL}$ [5], MMP [16] and PROJ [1], are omitted for clarity of presentation, they did not in general perform better than the methods shown. Unlike our algorithms, all of the methods we compare against use knowledge of the environment and consequently enjoy an additional advantage.

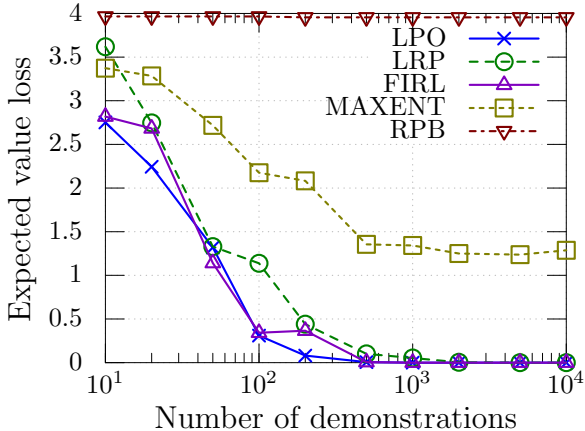

Figure 2: Gridworld results.

Figure 1(a) shows the expected loss $\ell$ for **blackjack** for LPO, LRP and prior algorithms. We can see that most algorithms found a good policy as the amount of data increases. However, RPB, MMP and MWAL (not shown) completely failed to find a reward inducing a policy close to the expert's one (always hits) even after $10^4$ demonstrations. The performance of MAX-ENT is better but does not improve with more data, in contrast to the original Bayesian approach[2] RPB, and the maximum a posteriori approach MAP$_{IRL}$. The best competing approach is FIRL, but LRP manages to surpass its performance, even without knowledge of the dynamics. In addition, both LPO and LRP require significantly less computation time than all other approaches, as can be seen in Fig. 1(b).

Figure 2 presents the expected value loss on the **gridworld**. The performance of LRP, LPO, and FIRL increases with more demonstrations. With less data, we can see that LPO slightly outperforms all other algorithms including FIRL. Again, the performance of MAX-ENT does not improve with more data. Also the performance of the original Bayesian approach RPB does not anymore increases with more data.

As a final example, we also present results on **tic-tac-toe** in Fig 3. This way, one may obtain an idea of how well our methods might be applicable to larger games. Figure 3(a) presents the value difference when the opponent is random opponent, while Figure 3(b) shows the results when the opponent is non deterministic but optimal. Against an optimal opponent, LPO, LRP and MAX-ENT quickly converge to the optimal (minimax) policy. Against a random opponent both LPO and LRP were able to outperform the optimal (minimax) expert while MAX-ENT did not improve from the expert. This can be explained by the fact that MAX-ENT assumes that the expert is optimal and then try to find a reward function which makes the expert optimal. In constrast, LPO and LRP considers a suboptimal expert ($\epsilon$-optimal expert), then try to find the true reward of this expert in order to outperform it if possible. Intuitively, the policy found by LPO and LRP takes a little risk in order to beat the expert against random opponents. At the same time, the found policy never looses against an optimal (minimax) expert. In this experiment, FIRL (not shown) completely failed to find a good policy. Its curve is y=1 for both the random or the optimal opponent.

---

[2]We also compared our method against the other models detailed in [7, 17], but RPB gave the best results in the tested domains.

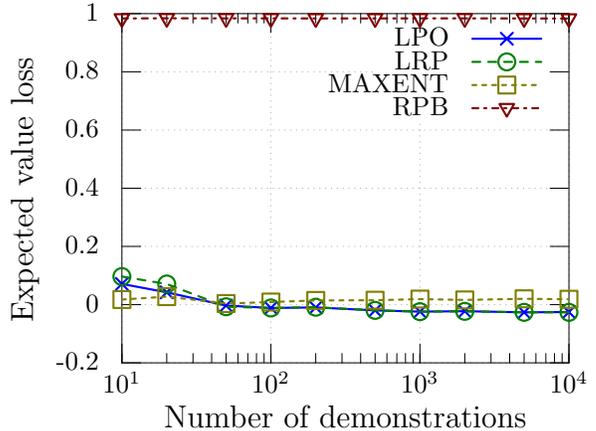

(a) Random opponent

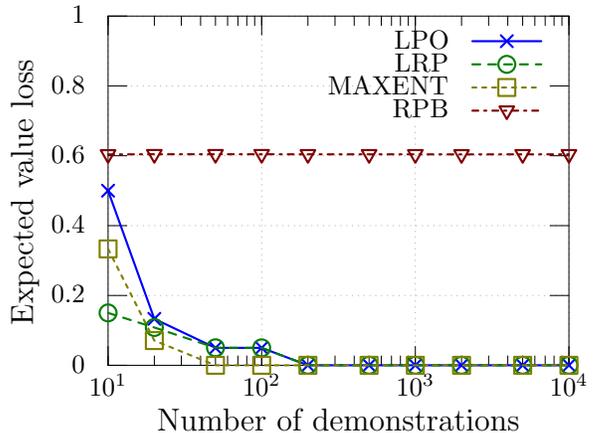

(b) Optimal opponent

Figure 3: Tic-Tac-Toe results

Overall, our algorithms are extremely robust and perform near-optimally in all domains. In the blackjack domain they reach the performance of the optimal policy and surpass all alternatives, all of which know the environment model *a priori*. In the gridworld, they compete well against FIRL and even outperformed it with less data in spite of the sophisticated feature discovery method and the knowledge of the dynamics by FIRL. Finally, the tic-tac-toe illustration demonstrates that such methods may also be useful in larger games.

Finally, it should be noted that our algorithms always outperform methods using a linear combination of the features. This could be explained by the fact that our cost function is convex and thus we are guaranteed to find the global solution. This is supported by our results for the fully Bayesian RPB method, as well as our preliminary results using different priors, which result in a non-convex problem.

## 6 Conclusion

We have provided a set of MAP approaches for probabilistic inverse reinforcement learning in unknown environments. These approaches are inspired by probabilistic models originally presented in [7, 17]. By avoiding full Bayesian inference, we were able to extend these methods to the case of unknown dynamics, Markov games and feature observations instead of actual state-action observations. In addition, we proposed an interesting simplification to the policy optimality model presented in [7], such that the set of all $\epsilon$-optimal reward functions for a particular policy does not need to be searched for the policy optimality algorithm.

Experimentally, both algorithms have a performance which equals and even surpasses that of algorithms which assume some knowledge of the underlying dynamics. This includes both approaches which use the dynamics analytically and approaches which only sample from the dynamics. In addition, the computational requirements of both methods are modest and significantly lower than those of the alternatives we tried.

One question is whether it is possible or useful to extend these algorithms to the nonlinear case. This can be done by using some kind of kernel, thus preserving the linearity in the features. This could be of interest for applications where the reward is complex with respect to the features. Finally, we would like to apply these methods to larger problems, and in particular to learning to play games from databases of expert play. It would be interesting to do so while extending them to multiple reward functions, such as for the models considered in [6, 7].

**Acknowledgements**

We wish to thank the anonymous reviewers for their comments on this as well as an earlier version of this paper. This work was partially supported by the Marie Curie Project ESDEMUU, Grant Number 237816.